\documentclass[letterpaper, 10 pt, conference]{ieeeconf}
\IEEEoverridecommandlockouts                              
\usepackage{amsmath,amssymb,amsfonts}
\usepackage{graphicx}
\usepackage{textcomp}
\usepackage[table]{xcolor}
\usepackage{booktabs}
\usepackage{makecell}
\usepackage{multirow}
\usepackage{caption}
\usepackage{subcaption}
\usepackage{algpseudocode}
\usepackage{algorithm}
\usepackage{float}
\usepackage[hidelinks]{hyperref}
\hypersetup{
    colorlinks,
    linkcolor={red!30!black},
    citecolor={blue!30!black},
    urlcolor={blue!30!black}
}
\usepackage{diagbox}
\usepackage{lineno}
\usepackage{setspace}
\usepackage{ulem}
\usepackage{xcolor}
\usepackage{moreverb,url}
\usepackage{amssymb}
\usepackage{adjustbox}
\usepackage{bm}
\newcommand{\SmallUpperCase}[1]{\textsc{\MakeLowercase{#1}}}  
\newcommand{\etal}[1]{et al. }  

\usepackage[capitalize]{cleveref}
\crefname{section}{Sec.}{Secs.}
\Crefname{section}{Section}{Sections}
\Crefname{table}{Table}{Tables}
\crefname{table}{Tab.}{Tabs.}

\title{\LARGE \bf
From Web Data to Real Fields: Low-Cost Unsupervised Domain Adaptation for Agricultural Robots
}

\author{Vasileios Tzouras$^{1}$, Lazaros Nalpantidis$^{1}$, and Ronja Güldenring$^{1, *}$%
\thanks{This work has been supported by the European Commission and European GNSS Agency through the project ``Galileo-assisted robot to tackle the weed Rumex obtusifolius and increase the profitability and sustainability of dairy farming (GALIRUMI)", H2020-SPACE-EGNSS-2019-870258.}%
\thanks{$^{1}$All authors are affiliated with Technical University of Denmark, Kongens Lyngby, Denmark
        {\tt\small \{s223182, lanalpa, ronjag\}@dtu.dk}}%
\thanks{$^{2}$This work has been submitted to the IEEE for possible publication. Copyright may be transferred without notice, after which this version may no longer be accessible.}
}

\begin{document}

\maketitle

\begin{abstract}
In precision agriculture, vision models often struggle with new, unseen fields where crops and weeds have been influenced by external factors, resulting in compositions and appearances that differ from the learned distribution. This paper aims to adapt to specific fields at low cost using Unsupervised Domain Adaptation (UDA). We explore a novel domain shift from a diverse, large pool of internet-sourced data to a small set of data collected by a robot at specific locations, minimizing the need for extensive on-field data collection. Additionally, we introduce a novel module -- the Multi-level Attention-based Adversarial Discriminator (MAAD) -- which can be integrated at the feature extractor level of any detection model. In this study, we incorporate MAAD with CenterNet to simultaneously detect leaf, stem, and vein instances. Our results show significant performance improvements in the unlabeled target domain compared to baseline models, with a 7.5\% increase in object detection accuracy and a 5.1\% improvement in keypoint detection.
\footnote[3]{Our code will be made publicly available on acceptance.}
\end{abstract}

\section{Introduction}\label{sec:intro}

Deploying vision-guided robots to perform agricultural tasks can reduce the need for manual labor and use of chemicals~\cite{Azghadi2024PreciseRW, Blasco2002AEAutomationAE, Wu2020RoboticWC} and ultimately increase agriculture's resilience to climate change~\cite{Siddiqui2021WeedMA}. Unfortunately, such robotic systems require large amounts of annotated data to train their perception systems, which is costly and infeasible for every field condition and flora composition. 

\begin{figure}[t]
    \centering
    \includegraphics[width=0.9\linewidth]{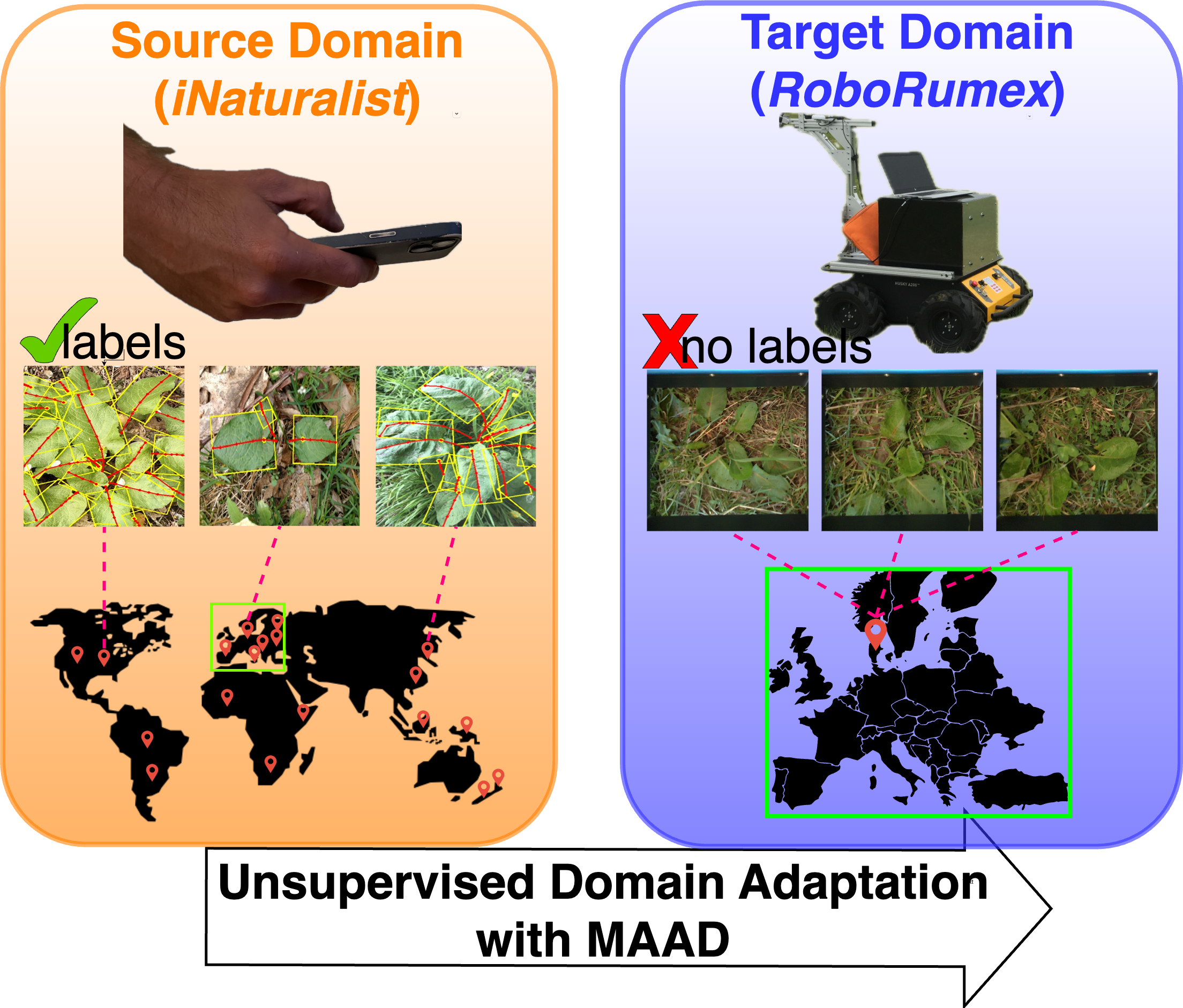} 
    \caption{We explore a novel domain shift: from internet-sourced data from all over the world (source domain) to images that have been collected with an agricultural robot platform at very specific locations (target domain). The source domain provides a large number of images with a wide variety of backgrounds and plant appearances. The data can be easily obtained without relying on agricultural vehicles and growing seasons. Finally, we introduce a Multi-level Attention-based Adversarial Discriminator (MAAD) to adapt the detection model to the target domain, which is represented by a small set of images collected at a specific location and with fixed camera settings.}
    \label{fig:intro}
\end{figure}

To partially alleviate this challenge, Domain Adaptation (DA) techniques in agriculture typically investigate domain shifts from one field to another~\cite{Gogoll2020UnsupervisedDA, Magistri2023FromOF}. However, nowadays, the abundance of online visual data from around the world---freely available from providers such as iNaturalist---opens up new possibilities at practically no data acquisition cost. In this work, we explore a novel domain shift from internet-sourced plant data to a small set of images collected at a specific field location under fixed camera settings, as depicted in Fig.~\ref{fig:intro}. Using the source domain, a powerful base model can be trained with large amounts of diverse data, whereas during the field adaptation phase, the base model can be adapted with a small set of freshly collected images to the unique characteristics of the field through DA.
More specifically, we propose a novel approach using Unsupervised Domain Adaptation (UDA) to transfer a model trained on web-sourced images to real-world conditions, specifically applying it to the RumexLeaves dataset~\cite{Gldenring2023ZoomIO}. Our method introduces a Multi-level Attention-based Adversarial Discriminator (MAAD) designed to force the model to learn feature distributions consistent across different domains. MAAD incorporates two adversarial discriminators operating at low and high feature levels, distinguishing between source and target domain inputs, while a primary object detection network generates detection predictions. Before feeding feature maps to the discriminators, we add a spatial attention module to focus on relevant regions. Experimental results on the RumexLeaves dataset demonstrate significant performance improvements in the unlabeled target domain compared to other baseline models used, confirming the effectiveness of our approach.

Our contributions can be summarized as follows: (i) we explore a novel domain shift for agricultural robotics from internet-sourced plant data to a specific field, rather than using the common field-to-field approach. More concretely, our study uses the publicly available RumexLeaves dataset~\cite{Gldenring2023ZoomIO} to simultaneously detect leaf instances with oriented bounding boxes and stem and vein lines with keypoints. The performance is evaluated for a range of well-established UDA methods such as CycleGAN~\cite{Zhu2017UnpairedIT}, MMD~\cite{Gretton2012AKT}, DANN~\cite{Ganin2015DomainAdversarialTO}, and MAAD; (ii) We propose a novel module, MAAD, that prioritizes informative regions at multiple levels to improve feature transfer between domains. MAAD demonstrates improvements over the baseline with a 7.5\% increase of leaf instance detection and 5.1\% stem and vein instance detection. Especially in the task of keypoint detection, it outperforms the alternative UDA methods.
\section{Related Work}
\textbf{Unsupervised Domain Adaptation (UDA).} Unsupervised Domain Adaptation (UDA) aims to generalize a model trained on a labeled source domain to a related target domain without labeled samples. Distance metrics are often used to measure the differences in distributions between source and target domains, with the goal of minimizing this difference through network training~\cite{Sun2016DeepCC, Tzeng2014DeepDC}. Adversarial learning methods, inspired by GANs, have become popular in UDA. These methods train both a domain classifier and a feature extractor simultaneously or use a Gradient Reversal Layer (GRL) to induce domain confusion by reversing gradients during backpropagation~\cite{Ganin2015DomainAdversarialTO, Hsu2020EveryPM, Saito2018StrongWeakDA}. On the other hand, image-to-image translation uses GANs~\cite{Hoffman2017CyCADACA, Inoue2018CrossDomainWO, Zhu2017UnpairedIT} or traditional image processing techniques~\cite{Abramov2020KeepIS, Yang2020FDAFD} to transform images from the source domain to an intermediate domain that matches the visual characteristics of the target domain. Another category of UDA methods is based on pseudo-labeling, which uses unlabeled target domain data by generating pseudo-labels from confident predictions, thereby improving model robustness through self-supervised learning~\cite{Mattolin2022ConfMixUD, RoyChowdhury2019AutomaticAO}. Finally, in the teacher-student framework, the teacher model, which is an average of the student model's past states, provides consistent pseudo-labels to guide the student model's training on the target domain~\cite{Ramamonjison2021SimRODAS, Zhou2022SSDAYOLOSD}.

\textbf{Adversarial Learning.} Adversarial learning methods are among the most common approaches used to tackle domain shift problems, either on their own or in combination with the other methods mentioned above. Domain-Adversarial Neural Network (DANN), introduced by Ganin \etal~\cite{Ganin2015DomainAdversarialTO}, improves feature separability for the primary task and robustness against domain shifts by using a domain-adversarial training framework that includes a GRL and a domain classifier to learn domain-invariant features. Chen \etal~\cite{Chen2018DomainAF} were among the first ones to introduce adversarial learning to address domain shifts in object detection. Their method uses domain classifiers for image-level and instance-level shifts, improving them through strong consistency regularization to achieve a domain-invariant Region Proposal Network. Guan \etal~\cite{Guan2021UncertaintyAwareUD} dynamically adjust adversarial learning strength based on sample alignment entropy. They used GRLs and domain classifiers to ensure feature extraction remains domain-invariant. On the other hand, Saito \etal~\cite{Saito2018StrongWeakDA} proposed cross-domain feature alignment using both global and local techniques. They employed two domain classifiers to minimize domain differences and integrated context vector-based regularization to improve stability and detection accuracy. Furthermore, Hsu \etal~\cite{Hsu2020EveryPM} used two domain discriminators to align features globally and with respect to the object centers.
\begin{figure*}[!htb]
    \vspace{2mm}
    \centering
    \includegraphics[width=0.9\linewidth]{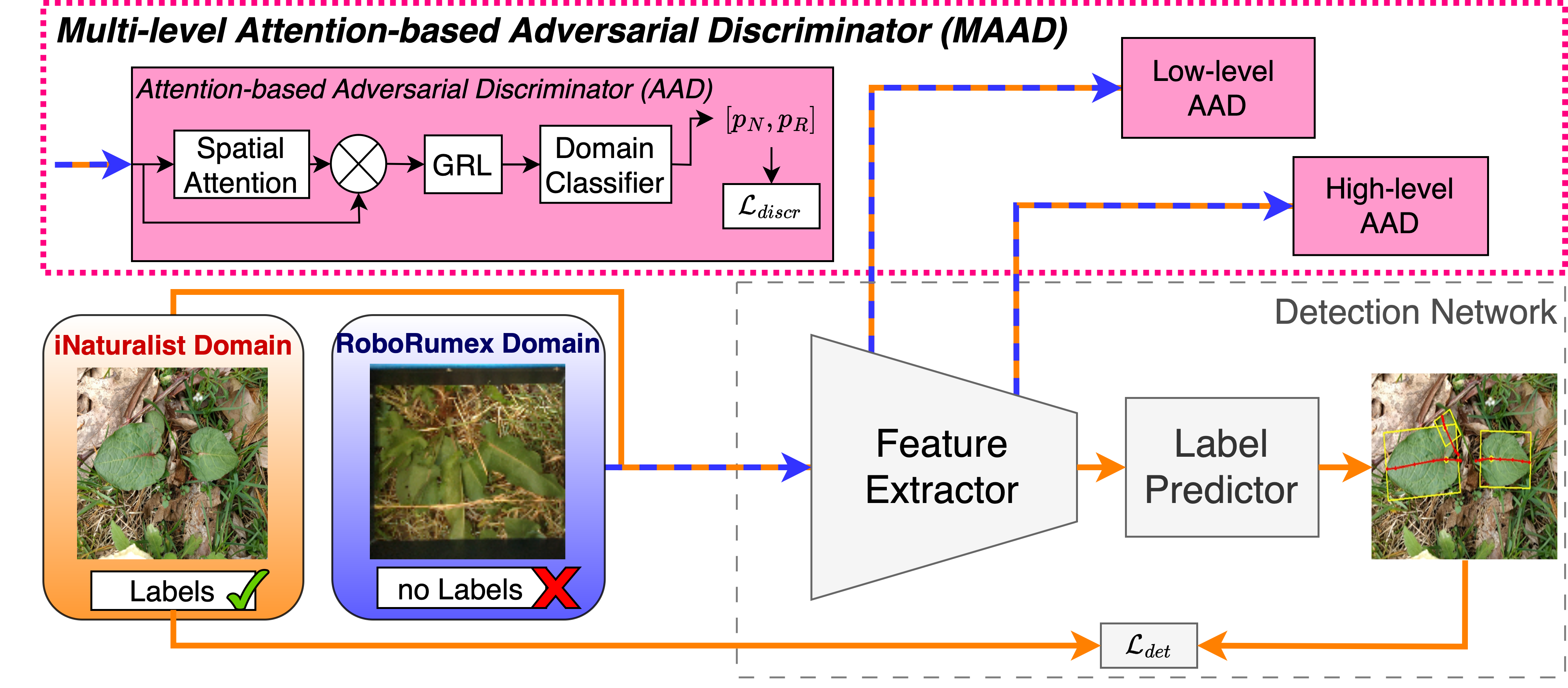}
    \caption{Our proposed Multi-level Attention-based Adversarial Discriminator (MAAD), shown in pink, is integrated with the feature extractor of the detection model. While the detection model is trained on source data only, MAAD processes both low- and high-level features of the source (orange) and target (blue) domains through the low-level AAD (LAAD) and high-level AAD (HAAD). The Attention-based Adversarial Discriminator (AAD) applies spatial attention to focus on informative regions, followed by a Gradient Reversal Layer (GRL) and a domain classifier to differentiate between the two domains. MAAD enforces the feature extractor to learn domain-invariant features, which are then processed to predict labels.}
    \label{fig:network}
\end{figure*}

\textbf{UDA in Agriculture.} Several studies have addressed the problem of domain shift in agriculture, particularly focusing on tasks like semantic segmentation and classification. Wu \etal~\cite{Wu2023FromLT} addressed this problem in plant disease classification using an auxiliary uncertainty regularization term without the need to use adversarial learning approaches. 
Zhang \etal~\cite{Zhang2022UnsupervisedAD} applied a UDA method to extract agricultural land from remote sensing images. They employed a GAN with a transformer backbone and a multi-scale feature fusion module to handle different spatial resolutions. Furthermore, Gogoll \etal~\cite{Gogoll2020UnsupervisedDA} extended CycleGAN for plant classification, preserving semantic details during image transformation. Bertoglio \etal~\cite{Bertoglio2023ACS} further improved this approach by minimizing phase differences between real and transformed images.
Ilyas \etal~\cite{Ilyas2023OvercomingFV} used adversarial feature alignment and entropy minimization, along with a strategy that progressively increases data augmentation frequency during training, for UDA in crop-weed segmentation. In a different approach, Vasconcelos \etal~\cite{Vasconcelos2021LowCostDA} introduced a simpler method for weed segmentation by aligning the low-frequency spectrum information of source images with that of target images using the Fast Fourier Transform. Finally, Magistri \etal~\cite{Magistri2023FromOF} proposed a method for semantic segmentation across various fields without the need for manual annotations. They employed GANs to synthesize images integrating source domain content with target domain style. 

Previous approaches work with source and target domains that are similar in terms of sensor characteristics because the data has been collected with comparable sensor setups such as viewpoint, resolution, and post-processing. In our work, on the contrary, we aim to reduce the need for extensive on-field data collection. Our source domain consists of freely available, diverse internet-sourced images and is then adapted to one specific field captured with an agricultural platform. Furthermore, we perform one-stage UDA for a multi-task setting including leaf instance detection with oriented bounding boxes and stem and vein instance detection with keypoints.

\section{Method}
In this section, we present our proposed module MAAD in \cref{fig:network}, which is model-agnostic and can be integrated into any Convolutional Neural Network (CNN) at the feature extractor level. This flexibility makes it adaptable to a wide range of architectures and downstream tasks. Concretely, we demonstrate its capabilities within the popular CenterNet architecture~\cite{Zhou2019ObjectsAP} in a multi-task setting, simultaneously detecting leaf (as oriented bounding boxes) and vein/stem (as keypoints) instances.

\textbf{Problem Statement.} In this paper, we focus on UDA, where we aim to adapt a model trained on labeled data from a source domain $\mathcal{D}_s = \{(x^i_s, y^i_s)\}_{i=1}^{N_s}$ to an unlabeled target domain $\mathcal{D}_t = \{x^i_t\}_{i=1}^{N_t}$, where $x^i_s \in \mathcal{X}_s$ are input data with corresponding annotations $y^i_s \in \mathcal{Y}_s$. $N_s$ is the number of samples in the source domain, and $\mathcal{X}_s$, $\mathcal{Y}_s$ are the feature and label spaces in the source domain, respectively. $x^i_t \in \mathcal{X}_t$ represents the input images in the target domain $\mathcal{X}_t$, and $N_t$ refers to the number of samples in the target domain.

\subsection{Multi-level Attention-based Adversarial Discriminator (MAAD)}
Inspired the DANN framework~\cite{Ganin2015DomainAdversarialTO}, we propose a Multi-level Attention-based Adversarial Discriminator (MAAD), where two Attention-based Adversarial Discriminators (AAD) are integrated with the feature extractor of the detection network at different levels. The High-level AAD (HAAD) aligns high-level features (e.g. plant appearance, lighting conditions) at the last block of the feature extractor, and the Low-level AAD (LAAD) aligns low-level features (e.g., edges, textures) at the first residual block of the backbone network. By integrating both discriminators, the model achieves alignment across different feature hierarchies, ensuring that both low-level texture information and high-level semantic information are aligned between the source and target domains.
MAAD operates within a minimax game framework, aiming to minimize detection loss on labeled source data using the main detector network while simultaneously maximizing the loss of the two AADs, which is equivalent to minimizing their negative loss: 
\begin{equation}
    \operatorname*{min}_{G_f} \operatorname*{max}_{G_{d_i}} \mathcal{L} = \mathcal{L}_{det} + \lambda_{\text{\SmallUpperCase{HAD}}} \mathcal{L}_{discr}^{\text{\SmallUpperCase{HAAD}}} + \lambda_{\text{\SmallUpperCase{LAD}}}\mathcal{L}_{discr}^{\text{\SmallUpperCase{LAAD}}}
    \label{eq:centda_obj}
\end{equation}
where $\mathcal{L}_{det}$ is the overall detection loss (\cref{eq:centernet}), $\mathcal{L}_{discr}^{\text{\SmallUpperCase{HAAD}}}$ is the HAAD loss (\cref{eq:BCE}), $\mathcal{L}_{discr}^{\text{\SmallUpperCase{LAAD}}}$ is the LAAD loss (\cref{eq:least_squares}), $\lambda_{\text{\SmallUpperCase{HAD}}}$ and $\lambda_{\text{\SmallUpperCase{LAD}}}$ are trade-off AD weights that control the balance between detection and AAD losses, $G_f$ is the feature extractor and $G_{d_i}$ are the domain classifiers.
\\\\
The Attention-based Adversarial Discriminator (AAD) consists of three components: the domain classifier to distinguish between the source and target domains, a GRL to enforce domain-invariant feature learning, and a spatial attention module to highlight key features during training. All components are explained in more detail below.

\textbf{Domain classifiers.} 
We designed two different classifiers for the different types of features that need to be aligned.

For the HAAD, we design the classifier with fully convolutional layers with 4$\times$4 filters, containing 128, 256, 512, and 1 filters per layer. Stride and padding adjustments are made to control spatial resolutions. Leaky ReLU activation with a slope of $\alpha = 0.2$ is applied in each layer, except the last. To stabilize training, we use batch normalization after the second and third layers. We train HAAD using Binary Cross-Entropy (BCE) loss, following other approaches~\cite{Ganin2015DomainAdversarialTO, Hsu2019ProgressiveDA, Saito2018StrongWeakDA}. The BCE loss is defined as:
\begin{equation}
   \mathcal{L}_{\text{discr}}^{\text{\SmallUpperCase{HAAD}}} = -\sum_{i, h, w}  d_i \log \hat{p}_i^{(h, w)} + (1 - d_i) \log(1 - \hat{p}_i^{(h, w)})
\label{eq:BCE}
\end{equation}
where $\hat{p}$ is the predicted distribution, $d$ is the domain class, $h$ and $w$ are the height and width of the image, respectively, of the $i^{th}$ training image.

For the LAAD, we adapt the HAAD architecture to focus on low-level features by using a 1$\times$1 kernel size, a stride of 1, and zero padding. Instead of BCE loss as in HAAD, we apply Least Square loss to better align low-level features, following previous studies~\cite{Mao2016LeastSG, Saito2018StrongWeakDA}:
\begin{equation}
   \mathcal{L}_{\text{discr}}^{\text{\SmallUpperCase{LAAD}}} = \sum_{i, h, w} d_i (\hat{p}_i^{(h, w)} - 1)^{2} + (1 - d_i) (\hat{p}_i^{(h, w)})^{2} 
   \label{eq:least_squares}
\end{equation}
where $\hat{p}$ is the predicted distribution, $d$ is the domain class, and $h, w$ are the image dimensions of the $i^{th}$ training image.

\textbf{Gradient Reversal Layer (GRL).} 
To align feature distributions across domains and produce domain-invariant features, we insert a Gradient Reversal Layer (GRL) before the domain classifier. The GRL acts as an identity function during forward propagation. During backpropagation, it reverses gradients by multiplying them with $-\lambda_p$ to maximize domain confusion.

\textbf{Spatial Attention.} 
To overcome the limitations of traditional adversarial DA methods, which align entire image representations without accounting for regional differences, we propose a spatial attention module inspired by Woo \etal~\cite{Woo2018CBAMCB} that improves adaptation by focusing on the most informative areas, thus enhancing the model's ability to distinguish between the two domains.
The spatial attention map $M \in \mathbb{R}^{H \times W}$, where $H$ and $W$ are the height and width of the feature map, is computed using the outputs from either the first residual block or the final block of the feature extractor, providing the flexibility to capture both low-level and high-level features. Given an input feature map $F \in \mathbb{R}^{C \times H \times W}$, the module aggregates spatial information by applying average and max pooling along the channel dimension, producing two 2D maps, $F_{\text{avg}} \in \mathbb{R}^{1 \times H \times W}$ and $F_{\text{max}} \in \mathbb{R}^{1 \times H \times W}$. These feature maps are concatenated to create a feature descriptor, which is then passed through a $7 \times 7$ convolutional layer to learn spatial dependencies, followed by a Sigmoid activation to generate $M$ (\cref{eq:attention}). The attention map $M$ applies a Hadamard product to the original feature map $F$, resulting in attention-weighted features $F' = M \odot F$, which are then passed through the GRL.

\begin{equation}
    M(F) = 
    \sigma \left( f \left( \left[ F_{\text{avg}}; F_{\text{max}} \right] \right) \right)
    \label{eq:attention}
\end{equation}
where $\sigma$ is the Sigmoid function, $f$ the convolutional layer, and $F_{\text{avg}}$ and $F_{\text{max}}$ are the feature maps obtained from average and max pooling, respectively.

\subsection{CenterNet}
CenterNet by Zhou \etal~\cite{Zhou2019ObjectsAP} is an anchor-free, one-stage object detector that identifies objects by their center points, with additional properties regressed from these centers. After feature extraction, Deformable Convolutional (DC) layers~\cite{Zhu2018DeformableCV} enhance receptive fields by adjusting filter weights based on learned offsets. Because DC layers do not support seeding, we trained our models five times, reporting the mean and standard deviation.

We use the publicly available repository from Güldenring \etal~\cite{Gldenring2023ZoomIO}, a modified version of CenterNet. This repository performs both Oriented Bounding Box (OBB) and keypoint detection, utilizing ResNet50 for feature extraction. It includes three task-specific heads: one for OBB regression for each leaf instance, one for keypoint regression along leaf stems and veins, and one for classification to refine the keypoints. The objective function of CenterNet combines L1 loss, \(\mathcal{L}_{1}\), for regression tasks and focal loss, \(\mathcal{L}_{f}\), for heatmap tasks, weighted by their importance:
\begin{equation}
    \mathcal{L}_{det} = \lambda_{cp} \mathcal{L}_{f} + \lambda_{off} \mathcal{L}_{1} + \lambda_{kp} \mathcal{L}_{1} + \lambda_{kphm} \mathcal{L}_{\text{f}} + \lambda_{obb} \mathcal{L}_{1} 
    \label{eq:centernet}
\end{equation}
where $\lambda_{cp}$, $\lambda_{off}$, $\lambda_{kp}$, $\lambda_{kphm}$, and $\lambda_{obb}$ are weights for the centerpoint heatmap, offset regression, keypoint regression, keypoint refinement heatmap, and OBB regression, respectively.

\section{Experiments}
\textbf{Implementation Details.} We implemented our network using PyTorch on an NVIDIA Tesla V100 with 32 GB of memory. We trained our model for 5500 epochs with a batch size of 32 (16 per domain) and optimized it using the Adam optimizer. The learning rate was set to 5$\times$10\textsuperscript{-4} for the detection network and 1$\times$10\textsuperscript{-4} for the discriminators, with a weight decay of 1$\times$10\textsuperscript{-5}. After each milestone epoch (2000, 3500, and 4500), the learning rate was decreased by a factor of 0.5 using a multi-step scheduler. We initialized the network weights from a normal distribution $\mathcal{N}(0, 0.001)$ and used Kaiming normal initialization for the discriminators. Throughout our experiments, we resized all images to 512$\times$512 pixels. To enlarge the dataset, we applied various augmentation methods, including random flipping, zooming out, color jitter, 90$^{\circ}$ rotation, Gaussian noise, random adjustments to brightness and contrast, and image normalization.

\textbf{Baselines.} We used three models as benchmarks: Maximum Mean Discrepancy (MMD) with a Gaussian Radial Basis Function kernel to compare source and target domain distributions after CenterNet's final up-sampling block (with a weight of 0.001 to balance detection and MMD losses), a DANN-inspired~\cite{Ganin2015DomainAdversarialTO} model where a combination of GRL and domain classifier is applied at a high-feature level, and CycleGAN trained patch-wise with a patch size of 128$\times$128 as in~\cite{Klages2019PatchBasedGA, Yu2019UnsupervisedDA}.

\textbf{Dataset.} In our experiments, we used the RumexLeaves dataset published by Güldenring \etal~\cite{Gldenring2023ZoomIO}. This dataset contains 809 images with fine-grained annotations of \textit{Rumex obtusifolius L.} weeds, including 7747 annotations with pixel-level segmentation masks for each leaf instance, as well as polylines for stems and veins of leaves. The dataset contains images from two related domains: 690 images sourced from the plant publisher iNaturalist and 119 images---referred to as RoboRumex---collected with an Intel Realsense L515 sensor mounted on a Clearpath Husky UGV at three farms around Copenhagen, Denmark.

\begin{table*}[!tb]
    \vspace{1mm}
    \centering
    \caption{Quantitative differences between the source and target domains. The RoboRumex domain includes smaller plants with an overall smaller number of leaves and leaf sizes (\textit{width per image [\%]}, \textit{height per image [\%]}). Furthermore, the RoboRumex images appear darker (\textit{intensity, brightness}) with less background clutter and variety (\textit{avg. edge magnitude}).}
    \label{tab:dataset_info}
    \resizebox{1.0\linewidth}{!}{
    \begin{tabular}{@{}l|c|c|c|c|c|c|c@{}}
        \toprule
        \textbf{Domain} & \textbf{\# leaves} & \makecell{\textbf{\# leaves} \\  per image} & \makecell{\textbf{Leaf width} \\ per image [\%]} & \makecell{\textbf{Leaf height} \\ per image [\%]} & \textbf{Intensity} & \textbf{Brightness} & \textbf{\makecell{Avg. edge \\ magnitude}} \\ 
        \midrule  
        Source (iNaturalist) & 7166 & 10$\pm$8 & 39.00$\pm$23.00 & 36.00$\pm$20.00 & 105.65$\pm$20.28 & 0.46$\pm$0.08 & 0.34$\pm$0.14 \\ 
        Target (RoboRumex) & 581 & 5$\pm$2 & 28.00$\pm$11.00 & 17.00$\pm$6.00 & 63.40$\pm$6.80 & 0.29$\pm$0.03 & 0.19$\pm$0.02 \\ 
        \bottomrule
    \end{tabular}
    }
\end{table*}
\begin{figure*}[!tb]
    \centering
    \includegraphics[width=1.0\linewidth]{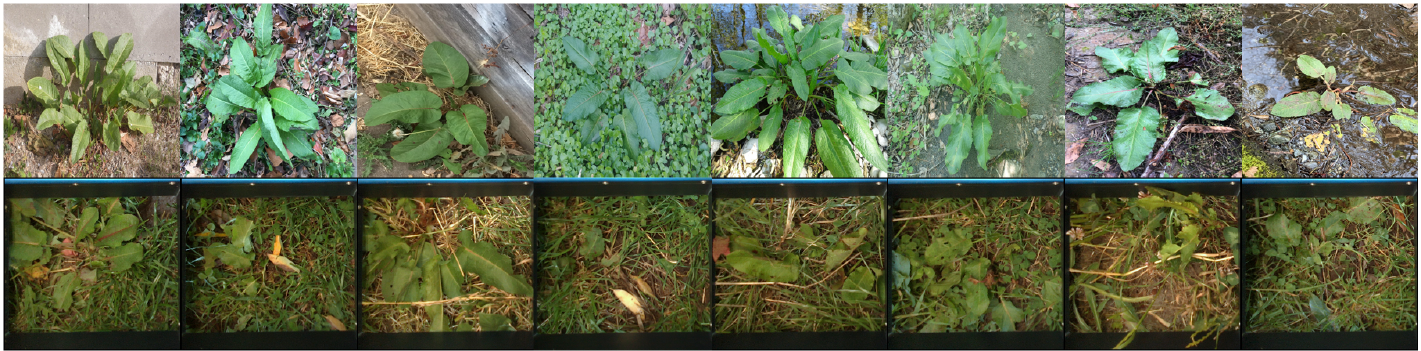} 
    \caption{Qualitative differences between the source (iNaturalist, top) and target (RoboRumex, bottom) domains. The iNaturalist images provide a variety of different background settings, such as mud, grass, dry leaves, stones, and a wide variety of plant sizes ranging from 2 to 25 leaves. On the contrary, RoboRumex data provides a more uniform appearance with plants of similar sizes and mainly grass as background.}
    \label{fig:qual_roborumex_vs_iNat}
\end{figure*}

These different image sources introduce an inherent domain shift within the dataset, as highlighted in ~\cref{tab:dataset_info}. For example, images from iNaturalist average approximately 10 leaves per image, while RoboRumex averages about 5 leaves per image. Additionally, bounding box size analysis reveals that the iNaturalist dataset has larger leaf dimensions compared to the smaller ones in RoboRumex. To measure the background characteristics, we extract the edges with the Sobel filter and report the average edge magnitude per image. For the RoboRumex domain, we can observe significantly lower edge length with a low variance. A possible explanation for this could be that in RoboRumex the background is mostly represented by grass and clover, which consists of small edges. On the contrary, in iNaturalist a variety of different background settings is available, ranging from mud, grass, dry leaves, stones, etc. as illustrated in \cref{fig:qual_roborumex_vs_iNat}.

Due to an incorrectly configured flag in our code, we accidentally swapped the validation and test split of the RoboRumex domain of the dataset during our experimental evaluation. Concretely, in our work, the RoboRumex validation split corresponds to the original test split from Güldenring \etal~\cite{Gldenring2023ZoomIO} and vice versa. According to Güldenring \etal~\cite{Gldenring2023ZoomIO}, the validation and test split have been randomly assigned and both correspond to 15\% of the total image pool. Therefore, we do not see a value in repeating the same experiments of approximately 1500 hours of training (on Nvidia Tesla V100) which corresponds to a relevant amount of computational cost.

\textbf{Evaluation Metrics.} We evaluate the performance of oriented object detection (mAP\textsubscript{50,\SmallUpperCase{OBB}}) using the commonly used mAP\textsubscript{50} metric from the DOTA dataset, which is based on the VOC implementation for horizontal bounding boxes. To evaluate the keypoint detection performance, we use Object Keypoint Similarity (OKS) and Projected Object Keypoint Similarity (POKS). OKS measures the similarity between predicted and ground truth positions, considering spatial proximity and overlap. POKS, introduced by Güldenring \etal~\cite{Gldenring2023ZoomIO}, extends OKS by including pseudo keypoints. It projects predicted keypoints onto neighboring line segments of the ground truth and then calculates OKS. We report four metrics for keypoint detection: mAP\textsubscript{50:95,\SmallUpperCase{OKS}} for OKS and mAP\textsubscript{50:95,\SmallUpperCase{POKS}} for POKS concerning stem, vein, and all keypoints.

\section{Results}
\subsection{Quantitative Results}
In this section, we show the performance of our proposed method MAAD compared to other relevant baselines in~\cref{tab:final_results}.
\begin{table*}[tb!]
    \vspace{1mm}
    \centering
    \caption{Comparison of different UDA methods for the domain shift from iNaturalist to RoboRumex in the RumexLeaves dataset~\cite{Gldenring2023ZoomIO}. Our proposed method, MAAD, demonstrates substantial improvements in both OBB and keypoint detection compared to the baseline model, approaching oracle performance.
    }
    \label{tab:final_results}
    \begin{adjustbox}{width=0.6\linewidth} 
    \begin{tabular}{@{}l|ccc|c|c@{}}
        \toprule
        \textbf{Method} & \multicolumn{3}{c|}{\textbf{mAP\textsubscript{50:95,\SmallUpperCase{POKS}}}} & \textbf{mAP\textsubscript{50:95,\SmallUpperCase{OKS}}} & \textbf{mAP\textsubscript{50,\SmallUpperCase{OBB}}} \\ 
        & All & Stem & Vein & & \\ 
        \midrule  
         No UDA & 27.15$\pm$1.90 & 17.10$\pm$2.70 & 38.85$\pm$1.15 & 7.75$\pm$1.60 & 64.35$\pm$2.85 \\
         Oracle & 34.25$\pm$1.55 & 26.25$\pm$2.00 & 44.30$\pm$1.95 & 13.60$\pm$1.15 & 74.20$\pm$0.90 \\
         \midrule
         MMD~\cite{Gretton2012AKT} & \underline{30.75$\pm$1.35} & 18.40$\pm$1.65 & \underline{43.95$\pm$1.80} & 12.15$\pm$0.10 & 68.80$\pm$2.25 \\ 
         CycleGAN~\cite{Zhu2017UnpairedIT} & 29.55$\pm$0.55 & \underline{18.55$\pm$1.40} & 43.55$\pm$1.50 & \underline{12.80$\pm$1.70} & \textbf{75.55\bm{$\pm$}2.30} \\
         DANN~\cite{Ganin2015DomainAdversarialTO} & 27.90$\pm$1.45 & 17.30$\pm$0.65 & 42.20$\pm$2.00 & 10.00$ \pm$1.10 & 69.50$\pm$3.00 \\
        \midrule
        Ours (MAAD) & \textbf{32.25\bm{$\pm$}1.05} & \textbf{20.50\bm{$\pm$}2.55} & \textbf{44.45\bm{$\pm$}1.30} & \textbf{12.85\bm{$\pm$}1.35} & \underline{71.85$ \pm$2.25} \\
        \bottomrule
    \end{tabular}
    \end{adjustbox}
\end{table*}
As a reference point, we first reproduce the results of Güldenring \etal~\cite{Gldenring2023ZoomIO}, where no UDA is applied, but the model is trained only on iNaturalist datapoints. Note, that the results differ slightly from the results in~\cite{Gldenring2023ZoomIO} because the usage of DC does not allow for proper seeding. Another reference point is our oracle model, which is trained on RoboRumex datapoints only, i.e. it represents the best performance achievable with supervised learning. Finally, we apply commonly used UDA methods such as CycleGAN~\cite{Zhu2017UnpairedIT}, MMD with the Gaussian kernel~\cite{Gretton2012AKT}, and DANN~\cite{Ganin2015DomainAdversarialTO}.
\Cref{tab:final_results} shows that there is potential for adapting from internet-sourced images (i.e., iNaturalist) to specific fields (i.e., RoboRumex). For all the presented UDA methods, we see performance improvements over the model with no UDA. The best model MAAD shows a performance increase by 7.5\% in OBB detection and 5.1\% in keypoint detection, closing the gap with the oracle model to around 2\% and less than 1\%, respectively, demonstrating results comparable to supervised learning. We observed the smallest improvement in mAP\textsubscript{50:95,\SmallUpperCase{POKS}} for stem detection, with an increase of 3.4\%, compared to a 5.6\% improvement for vein detection. This difference is likely due to the greater visibility of veins over stems in cluttered environments.
For CycleGAN, we can see a particularly strong performance boost in OBB detection, even outperforming the oracle model by 1.35\%. Further investigation of style-transfer techniques could potentially improve our approach and lead to even better results.
Compared to DANN~\cite{Ganin2015DomainAdversarialTO}, MAAD outperforms it in all categories with improvements of 2.85\% for keypoint detection and 2.35\% for OBB detection. This highlights how important it is to also align low-level features and include attention modules in our model. Additionally, a notable 3.2\% improvement in stem detection highlights our model's robust performance in complex, cluttered environments.

\subsection{Ablation Studies}
\subsubsection{Impact of different components in MAAD} 
In~\cref{tab:combinations}, we present the impact of different design choices in our proposed approach MAAD. We show the performance of individual components as well as their combinations. 
Initially, we show the performance of the two discriminators High-level Adversarial Discriminator (HAD) and Low-level Adversarial Discriminator (LAD) individually, both with and without GRL. Without GRL, the discriminator applies standard GAN-based adversarial learning. The overall tendency is, that the introduction of GRL leads to improved adaption results. Combining the HAD with GRL in (B) results in improved performance for both detection tasks compared to (A), where no GRL is applied. For LAD in (C) and (D) improvements due to GRL can only be observed for the keypoint detection of 2.8\%, and 
a performance decrease of 2.4\% is shown for the OBB detection. In further experiments, we prioritize configurations that incorporate the discriminators with GRLs as they provide better overall multi-task performance for our subsequent experiments.
When comparing the performance of the high-level discriminator to the performance of the low-level discriminator, we can observe that  HAD+GRL (B) outperforms LAD+GRL (D) in mAP\textsubscript{50,\SmallUpperCase{OBB}} by approximately 3\%, likely due to better alignment of high-level features. However, for keypoint detection, we see a decrease of 1.15\% compared to LAD+GRL (D), suggesting that aligning low-level features can improve keypoint detection performance. 
By combining both discriminators in (E), we achieve improved performance in keypoint detection, though we compromise a bit of performance in OBB detection.
Integrating the spatial attention layer A along with GRL in (F) resulted in a 3.8\% improvement in OBB detection performance. Meanwhile, we observed that the keypoint detection performance was not influenced significantly by the addition of the attention module, remaining at 14.55\%. The addition of the spatial attention module proved to be beneficial, especially in terms of OBB detection, as it helps our model focus more on spatial features that are important for understanding the overall context of weed leaves.
\begin{table}[htb]
 \centering
    \caption{Ablation study on the different components of our proposed method MAAD. We gradually introduce different components to increase complexity: High-level Adversarial Discriminator (HAD), Low-level Adversarial Discriminator (LAD), Gradient Reversal Layer (GRL), and spatial attention (A). We show the performance for both keypoint detection, mAP\textsubscript{50:95,\SmallUpperCase{OKS}}, and OBB detection, mAP\textsubscript{50,\SmallUpperCase{OBB}}.}
    \label{tab:combinations}
    \resizebox{0.8\columnwidth}{!}{
    \begin{tabular}{@{}l|cccc|c|cc@{}}
        \toprule
        \textbf{ID} & \textbf{HAD} & \textbf{LAD} & \textbf{GRL} & \textbf{A} & \textbf{mAP\textsubscript{50:95,\SmallUpperCase{OKS}}} & \textbf{mAP\textsubscript{50,\SmallUpperCase{OBB}}} \\ 
        \midrule
        A & \checkmark & & & & 12.85 & 61.00 \\
        B & \checkmark & & \checkmark & & 13.25 & 69.25 \\
        C & & \checkmark & & & 11.60 & 68.65 \\
        D & & \checkmark & \checkmark & & 14.40 & 66.25 \\
        E & \checkmark & \checkmark & \checkmark & & \textbf{14.75} & 68.30\\ 
        F & \checkmark & \checkmark & \checkmark & \checkmark & \underline{14.55} & \textbf{72.10} \\
        \bottomrule
    \end{tabular}
    }
\end{table}
\subsubsection{Impact of AD Weight} 
In this section, we emphasize the critical importance of balancing detection and adversarial losses during optimization. We use $\lambda_{\text{\SmallUpperCase{HAD}}}$ and $\lambda_{\text{\SmallUpperCase{LAD}}}$ to control the trade-off between these losses. We conducted an ablation study on the RoboRumex validation split, varying $\lambda_{\text{\SmallUpperCase{HAD}}}$ from 0.001 to 1 and $\lambda_{\text{\SmallUpperCase{LAD}}}$ from 0.0001 to 0.01. In the ablation study, lower weights are assigned to low-level features in LAD due to their lesser impact on high-level task-specific representations. This allows a contribution of low-level features DA, while their influence is scaled in comparison to the more critical high-level features.
The results, presented in~\cref{tab:da_ablation}, indicate that very high $\lambda_{\text{\SmallUpperCase{HAD}}}$ values significantly reduce mAP, with the best performance achieved at $\lambda_{\text{\SmallUpperCase{HAD}}}$ = 0.001. Similarly, we observed that smaller $\lambda_{\text{\SmallUpperCase{LAD}}}$ values lead to better performance, and we set $\lambda_{\text{\SmallUpperCase{LAD}}}$ to 0.0001. 
\begin{table}[htb!]
    \centering
    \caption{Ablation study results on the impact of the AD weights, namely $\lambda_{\text{\SmallUpperCase{HAD}}}$ for the High-level Adversarial Discriminator (HAD) and $\lambda_{\text{\SmallUpperCase{LAD}}}$ for Low-level Adversarial Discriminator (LAD). We report the performance for the keypoint and OBB tasks.}
    \resizebox{0.8\columnwidth}{!}{
    \begin{tabular}{@{}l|c|c|cc@{}}
        \toprule
        \textbf{Block} & \textbf{AD Weight} & \textbf{mAP\textsubscript{50:95,\SmallUpperCase{OKS}}} & \textbf{mAP\textsubscript{50,\SmallUpperCase{OBB}}} \\ 
        \midrule  
         \multirow{4}{*}{HAD + GRL} & 1.0 & 9.10 & 63.25 \\ 
         & 0.1 & 12.60 & 65.70 \\
         & 0.01 & \textbf{13.40} & 61.45 \\
         & 0.001 & \underline{13.25} & \textbf{69.25} \\ 
        \midrule
        \multirow{3}{*}{LAD + GRL} & 0.01 & 6.40 & 64.85 \\ 
         & 0.001 & 6.70 & 66.70 \\ 
         & 0.0001 & \textbf{14.40} & \textbf{66.25} \\  
        \bottomrule
    \end{tabular}
     }
    \label{tab:da_ablation}
\end{table}
\begin{figure*}[htb!]
    \vspace{2mm}
    \centering
    
    \begin{subfigure}[b]{1.0\textwidth}
        \centering
        \includegraphics[width=\linewidth]{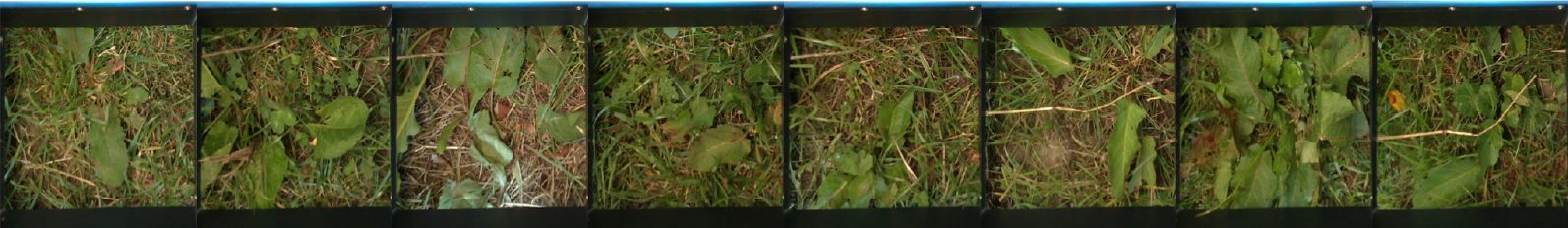}
        \caption{Input images.}
    \end{subfigure}
    
    \begin{subfigure}[b]{1.0\textwidth}
        \centering
        \includegraphics[width=\linewidth]{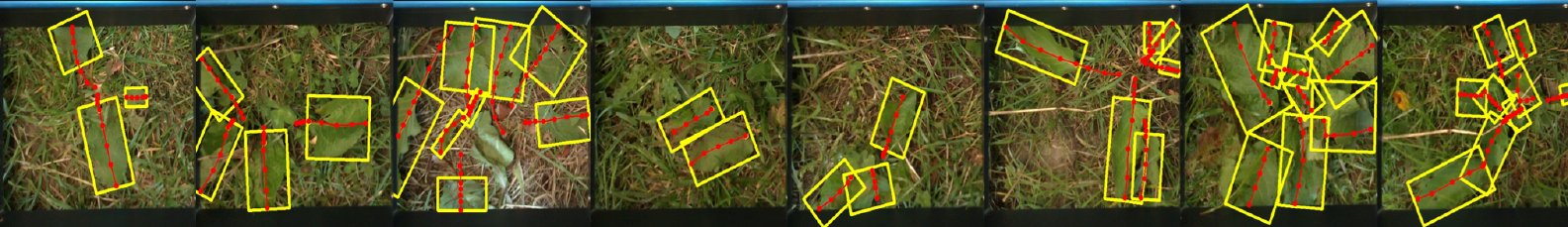}
        \caption{Ground truth.}
    \end{subfigure}
    
    \begin{subfigure}[b]{1.0\textwidth}
        \centering
        \includegraphics[width=\linewidth]{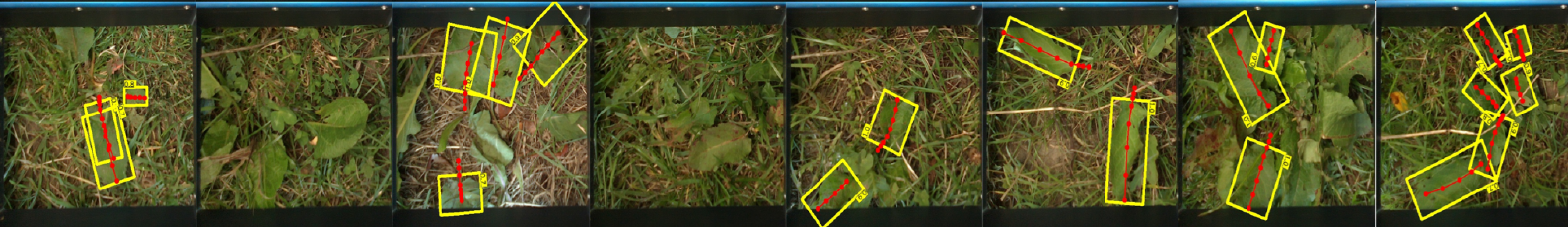}
        \caption{Predictions without UDA.}
    \end{subfigure}
    
    \begin{subfigure}[b]{1.0\textwidth}
        \centering
        \includegraphics[width=\linewidth]{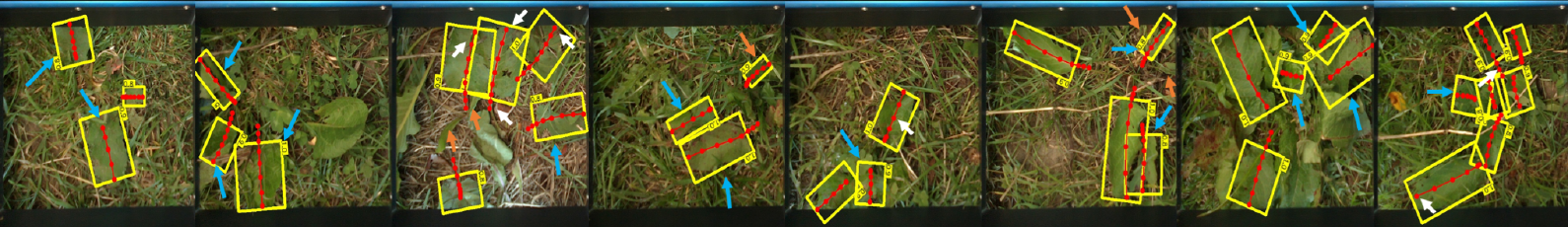}
        \caption{Predictions with UDA using our proposed method, MAAD.}
    \end{subfigure}
    
    \caption{Qualitative results on the RoboRumex test set: (a) input images, (b) ground truth, (c) predictions without UDA, and (d) predictions with UDA using our proposed MAAD. Blue and white arrows highlight improvements of MAAD compared to the baseline for OBB and keypoint detection, respectively. Orange arrows indicate FPs and small leaves that were not detected compared to the ground truth.}
    \label{fig:qualitative}
\end{figure*}

\subsection{Qualitative Results}
\Cref{fig:qualitative} illustrates some qualitative detection results on RoboRumex test split before (c) and after (d) implementing UDA with MAAD, along with the ground truth (b) and the original images (a). These results demonstrate that MAAD effectively identifies most weed leaves compared to the baseline model, as we observe differences marked with blue arrows for OBB detection and white arrows for keypoint detection. 
However, some False Positives (FP) persist, and the model occasionally misses very small leaves, as indicated by the orange arrows. We attribute the FPs to the complexity of the RoboRumex environment, characterized by more cluttered backgrounds mainly consisting of grass and clover which can sometimes resemble weed leaves in appearance and texture, potentially leading to misclassifications by the model. The problem of missed detections for very small leaves arises from the limited number of such samples in the training set, which restricts our model's ability to learn these characteristics. While we acknowledge that considering the use of DC to mitigate this challenge is valid, it may not have provided significant assistance in resolving the issue. Detecting very small objects is a well-known challenge in computer vision, as discussed by Li \etal~\cite{Li2022StepwiseDA}. 
\section{Conclusion}
In this paper, we investigate a novel domain shift for agricultural robotics, specifically from internet-sourced field data to images from fields at specific locations. We apply well-established UDA methods, such as MMD, CycleGAN, and DANN, to showcase the challenges posed by more difficult domain shifts. Furthermore, we introduce a Multi-level Attention-based Adversarial Discriminator (MAAD) that is integrated with the feature extractor of the detection model, processing both low- and high-level features while prioritizing informative regions through spatial attention. 
Experimental results demonstrate that MAAD significantly improves performance in the target domain, outperforming the baseline model (without UDA) with a 7.5\% increase in oriented bounding box detection and a 5.1\% increase in keypoint detection, achieving performance comparable to the oracle model.


\bibliographystyle{IEEEtran}
\bibliography{IEEEabrv, references}
\end{document}